\documentclass[10pt,twocolumn,letterpaper]{article}

\usepackage{cvpr}      
\usepackage{float}
\usepackage{dsfont}
\usepackage{amssymb}
\usepackage{tabularx}
\usepackage{duckuments}
\usepackage{microtype}
\usepackage{wrapfig}
\usepackage{makecell} 
\usepackage[dvipsnames]{xcolor}

\definecolor{cvprblue}{rgb}{0.21,0.49,0.74}
\usepackage[pagebackref,breaklinks,colorlinks,citecolor=cvprblue]{hyperref}
\usepackage{booktabs}
\usepackage{mathtools}

\def\paperID{6506} 
\def\confName{CVPR}
\def\confYear{2024}

\newcommand{\name}{Diffusion Handles\xspace}

\title{\name\\ Enabling 3D Edits for Diffusion Models by Lifting Activations to 3D}

\author{Karran Pandey\\
University of Toronto\\
{\tt\small karran@cs.toronto.edu}
\and
Paul Guerrero \\
Adobe Research\\
{\tt\small guerrero@adobe.com}
\and
Matheus Gadelha\\
Adobe Research\\
{\tt\small gadelha@adobe.com}
\and
Yannick Hold-Geoffroy\\
Adobe Research\\
{\tt\small holdgeof@adobe.com}
\and
Karan Singh\\
University of Toronto\\
{\tt\small karan@dgp.toronto.edu}
\and
Niloy Mitra\\
UCL, Adobe Research \\
{\tt\small n.mitra@cs.ucl.ac.uk}
}

\begin{document}
\twocolumn[{%
\renewcommand\twocolumn[1][]{#1}%
\maketitle

\begin{minipage}{\textwidth}
\vspace{-15pt}
\centering
\url{https://diffusionhandles.github.io}
\end{minipage}

\begin{center}
    \centering
    \captionsetup{type=figure}
    \includegraphics[width=\textwidth]{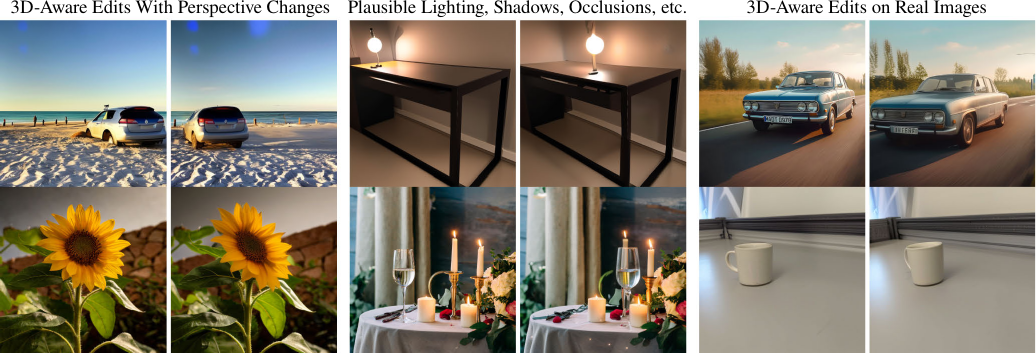}
    \captionof{figure}{\emph{Diffusion Handles} enable 3D-aware object edits (e.g., 3D translations and rotations), on images generated by diffusion models. Edited images exhibit plausible changes in 
    perspective, occlusion, lighting, shadow, and other 3D effects, without explicitly solving the inverse graphics problem. 
    Our method does not require training or fine-tuning and can be applied to real images through image inversion.}
    \label{fig:teaser}
\end{center}
}]

\begin{abstract}

\name is a novel approach to enabling 
3D object edits on diffusion images. We accomplish these edits using existing pre-trained diffusion models, and 2D image depth estimation,
without any fine-tuning or 3D object retrieval. The edited results remain plausible, photo-real, and preserve object identity. 
\name address a critically missing facet of generative image based creative design, and significantly advance the state-of-the-art in generative image editing. 
Our key insight is to lift diffusion activations for an object to 3D using a proxy depth, 3D-transform the depth and associated activations,
and project them back to image space. The diffusion process applied to the manipulated activations with identity control, produces plausible edited images showing complex 3D occlusion and lighting effects. 
We evaluate \name:  quantitatively, on a large synthetic data benchmark; and
qualitatively by a user study, showing our output to be more plausible, and better than prior art at both, 3D editing and identity control.

\end{abstract}

\section{Introduction}
\label{sec:intro}

\begin{figure}[t]
  \centering
    \includegraphics[width=\columnwidth]{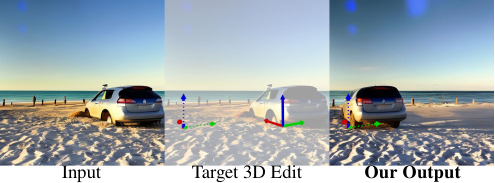} 
    \caption{\textbf{3D-aware image editing.} DiffusionHandles gives a user control over 3D transformations of 3D objects shown in images. We demonstrate 3D rotations and translations in our experiments. The original 3D pose of an object is visualized with a solid coordinate frame, and the target pose with a dotted coordinate frame.}
    \label{fig:intro}
\end{figure}

Text-to-image diffusion models~\cite{ramesh2022dalle,rombach2022latentdiffusion,saharia2022photorealistic} are the state-of-the-art in image generation. They produce photo-real outputs, effortlessly generate complex, high-resolution images, and support various forms of conditional generation~\cite{zhang2023_controlnet}. Pretrained diffusion models can be repurposed to support many image processing tasks~\cite{saharia2022palette}, such as, image in- or out-painting, superresolution, and denoising. 
However, there is limited support for object-centric editing in such images, where the 3D composition of scene objects can be changed, while preserving their identity.
Existing approaches treat such edits in image space, including: cutting and pasting objects to desired locations using object masks and regenerating the background~\cite{avrahami2022blended}; using gradient domain edits with some identity control~\cite{epstein2023diffusion}; or leveraging novel view synthesis with fine-tuned diffusion models~\cite{michel2023object} that are costly to train and can reduce model generality.
These approaches are particularly restrictive, for 3D object edits involving translations, rotations, and changes in scene perspective.
%
%
Moving the car to a new location on the beach in Figure~\ref{fig:teaser} for example, is non-trivial if the identity of the car has to be retained. The 3D edit should successfully capture complex light and shading effects, as well as a change in perspective (see Figure~\ref{fig:intro}), which is hard to achieve by enforcing object pixel intensity invariance used in image-space identity control.

We propose {\em \name} to support object-level edits aware of the underlying (hidden) 3D object structure. We demonstrate how to generate plausible
3D edits without solving the inverse graphics problem~\cite{interactiveimages,kholgade20143d} (i.e., without needing to recover the full scene geometry, materials, and illumination). In other words, we enable 3D-aware edits directly on 2D images generated by a diffusion model or actual photos that can be inverted~\cite{mokady2023null} into a diffusion model. At the heart of our method is a novel approach to {\em lift} diffusion activations to 3D, encoded as coarse proxy depth that can be estimated by a method (e.g.,~\cite{bhat2023zoedepth}). The lifted activations can then be moved or transformed in 3D scene space and projected back to the image plane using the estimated depth maps. 
We use these projected activation maps to guide the diffusion process to produce the final edited image (see Figure~\ref{fig:teaser}). Our approach is simple and effective, leverages pretrained diffusion models \emph{without fine-tuning or additional training data}, and produces plausible results even when the estimated depths have moderate warps.

Our method enables a range of 3D modifications to objects, such as translations that affect perspective, scaling, and some rotation. We accomplish this by converting the depth map into a point cloud and then applying the desired 3D transformation. To obtain this point cloud, we need the depth map of the object, either from a known template or by estimating it from the image. When working with a known template, precise alignment is necessary. When using an estimated depth, disocclusions 
can result in unknown depth regions, leading to uncertainty and loss of identity when the transformation is too extreme, especially with large rotations. 



We evaluate our method on various editing scenarios, including real and generated images. Since our approach allows for a new type of 3D-aware edits with diffusion-based generated images, we compare it to other editing methods capable of applying similar edits. Specifically, we compare against a 2D editing method using a similar activation-based guidance~\cite{epstein2023diffusion} and a 3D-aware editing method based on novel view synthesis that fine-tunes a diffusion model using 3D information~\cite{michel2023object}. In contrast, our method allows 3D-aware editing without the need for fine-tuning (see Table~\ref{tab:methodVariations}). We demonstrate the generalizability of our method on a large number of qualitative examples. Additionally, we conduct a user study to compare our results against existing baselines and ablated versions of our approach. 
In summary, to the best of our knowledge, we present the first editing framework that supports fine-grained 3D control over the object layout in diffusion images, without requiring any additional training.

\begin{table}[t]
\centering
\caption{
\textbf{Comparison to related methods.}
Our approach is unique in allowing 3D edits, without
any additional training, or 3D data.
Our Zero123 baseline allows for 2.5D edits
(denoted by *).
}
\footnotesize 
\renewcommand{\arraystretch}{1.1}
\setlength{\tabcolsep}{5pt}
\begin{tabularx}{\linewidth}{rccccc} 
\toprule
     & \makecell{3DIT\\\cite{michel2023object}} & \makecell{Diffusion \\ Self-Guid.\\\cite{epstein2023diffusion}} & \makecell{Zero123\\\cite{zero123}} & \makecell{Obj.Stitch\\\cite{song2023objectstitch}} & \makecell{Ours} \\
     \midrule
     Training-free? & $\times$ & \checkmark & $\times$ & $\times$ & \checkmark  \\
     No 3D data? & $\times$ & \checkmark & $\times$ & \checkmark & \checkmark \\
     3D edits? & \checkmark & $\times$ & \checkmark$^*$ & $\times$ & \checkmark \\
     \bottomrule
\end{tabularx}
\label{tab:methodVariations}
\end{table}

\section{Related Work}
\label{sec:related-work}

\paragraph{Text-Guided Image Generation.}
Seminal approaches for creating images from text prompts relied on a combination of image retrieval
and composition using user-created layouts~\cite{sketch2photo}.
Later, several representation learning techniques targeted creating joint representation for images
and text~\cite{mori1999,quattoni2007learning,srivastava2012}.
In the last few years, such multimodal approaches were scaled to hundreds of millions of text-image pairs by using modern deep learning architectures and contrastive learning~\cite{clip}.
Using those representations, initial attempts at image generation did so through a combination of gradient-based optimization
and image priors -- some hand-crafted~\cite{clipdraw}, others data-driven~\cite{styleclip,vqganclip}.
However, they suffered from slow runtime and had trouble generating visually appealing imagery.
These issues were addressed by several techniques that trained models for outputting images from
text prompts.
Such approaches relied on autoregressive~\cite{parti,cogview,chang2023muse} and diffusion models~\cite{ramesh2022dalle,nichol2021glide,makeascene}.
Follow-up works also investigated how to provide finer-grained control over the generative process (beyond
text prompts) like using regional prompting~\cite{multidiffusion,zeng2022scenecomposer} and additional user-provided image information like depth maps and edges~\cite{zhang2023_controlnet,t2i}.
Despite the photo-real quality of the generated images, those approaches do not allow users to manipulate existing image elements and, more importantly, provide any 3D-aware controls.
While our approach relies on existing diffusion models~\cite{rombach2022latentdiffusion}, we extend their capabilities
(without the need for any additional training) to allow users to manipulate objects in real or generated
images in a 3D-aware manner.

\paragraph{Image Editing with Generative Models.}
Generative models have been powering several image editing tasks, like inpainting~\cite{lugmayr2022repaint},
object insertion and harmonization~\cite{li2023image} and
stylization~\cite{wang2023stylediffusion}. 
For these traditional tasks, data-driven models offer a way to achieve superior control with
less user intervention.
They also enabled new tasks like synthesizing images from semantic segmentation maps~\cite{zhang2023_controlnet,park2019semantic,t2i} and text-annotated layouts~\cite{makeascene,spatext,multidiffusion,li2023gligen,chen2023training}.
More recently, open-ended text-guided image editing has been explored by combining large language models with
text-to-image generators~\cite{dalle3,instructpix2pix}.
Despite the impressive results, the previous methods do not allow users to preserve the appearance
of objects while manipulating the image elements.
This issue can be partially addressed by allowing users to edit images by dragging relevant keypoints~\cite{pan2023drag,mou2023dragondiffusion}.
Such controls are adequate for performing object deformation but might be cumbersome for other tasks like
changing object positioning in the scene.
Closely related to our work, Epstein \etal~\cite{epstein2023diffusion} address the problem of editing existing image elements
by manipulating the intermediate representation of text-to-image models.
They demonstrate how to alter individual object size and 2D position without resorting to any additional training.
Unfortunately, none of the aforementioned techniques target 3D object manipulation in images.
For that reason, they are incapable of addressing occlusions between edited entities in an image, their 3D position and out-of-plane
rotation, for example.
On the other hand, our approach is specifically designed to handle these scenarios.

\paragraph{3D-aware Image Editing.}
Several works have investigated editing 2D images in a 3D-aware manner by changing the viewpoint
in which a picture was taken.
This was initially attempted with hand-crafted priors~\cite{photopopup}, but was lately enhanced by data-driven
models performing a combination of inpainting and monocular depth estimation~\cite{kenburns,3dphotography,slide}.
While those techniques allow users to perform fine-grained camera motions, they do not
investigate how to manipulate particular objects in the scene.
Early attempts were not fully automatic but required
significant user input~\cite{interactiveimages} and existing 3D models~\cite{objmanipulation}.
More recent work relies in massive 3D object datasets~\cite{michel2023object} or in training regimes relying in image-based models
which hurt their interaction time~\cite{po2023compositional} or generality~\cite{BlockGAN2020}.
Zero123~\cite{zero123}, a notable recent method, does allow pseudo-3D rotations but requires access to 3D models to create a dataset in order to fine-tune a controlNet~\cite{zhang2023_controlnet} backend. 
In this work, we propose a technique that does not rely on any additional training, does not require large 3D datasets,
has an inference time similar to generating an image in text-to-image models, and allows editing
in-the-wild images without category-specific restrictions while maintaining realistic image quality.

\begin{figure*}[t]
    \centering
    \includegraphics[width=\textwidth]{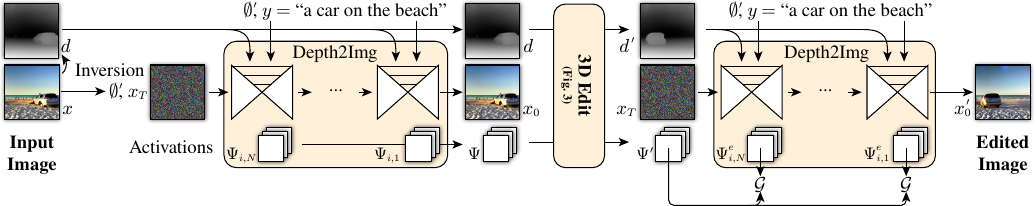}
    \caption{\textbf{Overview.} Starting from an input image $x$, we first estimate depth $d$ and invert the image into a depth-to-image diffusion model,
    giving us activations $\Psi$ that reconstruct the input image. A 3D transform supplied by a user can then be applied to $\Psi$ by lifting them to the 3D surfaces given by the depth map (Figure~\ref{fig:3dedit} shows details). Using the 3D transformed $\Psi'$ as guidance in a diffusion model allows us to generate an edited image that adheres to the edit, preserves the identity of the input image, and is plausible.}
    \label{fig:overview} 
\end{figure*}

\section{Overview}

Our goal is to imbue text-to-image generation pipelines with 3D-aware object edit handles, without requiring any fine-tuning of the generative model. In particular, given an image
of a 3D scene and a corresponding text prompt,
our method allows the user to perform a 3D transform,
such as translation, rotation and scale, on any object described in the text prompt.
Figure \ref{fig:teaser} shows several examples.

The given image is first inverted with a diffusion model, and an image of the \emph{edited} scene
is generated by the same diffusion model, guided by additional loss terms that we design to fulfill three tasks: (i)~to generate a 3D-transformed version of the edited object, (ii)~to otherwise preserve the appearance of all objects in the 3D scene, and (iii)~to still allow leveraging the prior of the diffusion model so that the edited object realistically interacts with its environment through lighting, shadows, etc.

We achieve these tasks by lifting activations of the diffusion model to the 3D surface of scene objects, where we can apply 3D transformations. The 3D-transformed activations can then serve as guidance when generating the edited image.
We
find that the strong prior of the diffusion model makes our method robust to inaccuracies and artifacts of the 3D surfaces we use in our edits, and that the approximate depth obtained from existing depth estimators is sufficient to allow for a wide range of 3D edits.

\section{Diffusion Models}

\paragraph{Training.}
During training, a fixed process adds a random amount of noise to an image $x$ to get a noisy image $\tilde{x}(t)$:
\begin{equation}
    \tilde{x}(t) = \sqrt{\alpha(t)}\ x + \sqrt{1-\alpha(t)}\ \epsilon,
\end{equation}
where $\epsilon \sim \mathcal{N}(\mathbf{0}, \mathbf{I})$ is Gaussian noise, and $t \in [0, T]$ parameterizes a noise schedule $\alpha$ that determines the amount of noise in $\tilde{x}(t)$, with $\alpha(0) = 1$ (no noise) and $\alpha(T) = 0$ (pure noise).
A denoiser $\epsilon_\theta$ with parameters $\theta$ is trained to predict the noise $\epsilon$ using the following loss:
\begin{equation}
    \mathcal{L_{\text{diff}}} = w(t) \|\epsilon_\theta(\tilde{x}(t); t, y, d) - \epsilon\|_2^2
\end{equation}
where  $d$ is a depth map, $y$ is an encoding of a text prompt, and $w(t)$ is a weighting scheme for different parameters $t$.
The parameter $t$ is sampled uniformly from $[0, T]$ in each training iteration.
Once $\epsilon_\theta$ is trained, $-\epsilon_\theta(\tilde{x}(t); t, y, d)$ defines a vector field in image space that points towards the natural (non-noisy) image manifold.

\paragraph{Inference.}
At inference time, an image is generated by starting from pure noise $\tilde{x}(T)$, and following the vector field $-\epsilon_\theta(\tilde{x}(t); t, y, d)$ towards the natural image manifold.
Multiple different samplers have been proposed~\cite{ho2020denoising, song2020denoising} to find a trajectory $x_T, x_{T-1}, \dots, x_0$
with a fixed number of $T$ discrete steps that starts at $x_T \coloneqq \tilde{x}(T)$ and ends in an image $x_0$ close to the natural image manifold.
Our method is compatible with any standard sampler; we describe the sampler we use in our experiments in the supplemental.

\paragraph{Guidance.}
The vector field $\epsilon_\theta(\tilde{x}(t); t, y, d)$ can be guided to minimize a custom energy $\mathcal{G}(\tilde{x}(t); t, y, d)$ by biasing each step with the gradient $\nabla_{\tilde{x}(t)}\ \mathcal{G}$ of the energy.
Apart from this form of guidance, most samplers also use classifier-free guidance~\cite{ho2022classifier} to more closely follow the text prompt $y$, by moving the vector $\epsilon_\theta(\tilde{x}(t); t, y, d)$ away from the vector $\epsilon_\theta(\tilde{x}(t); t, \emptyset, d)$ obtained with the encoding $\emptyset$ of an empty text prompt (also called \emph{null-text}). Similar to previous work~\cite{epstein2023diffusion}, we combine the two forms of guidance:
\begin{align}
\label{eq:guidance}
    \epsilon_\theta^\mathcal{G}(\tilde{x}(t); t, y, d)\ =\ & \ (1+\mu)\ \epsilon_\theta(\tilde{x}(t); t, y, d) \\
    & - \mu\ \epsilon_\theta(\tilde{x}(t); t, \emptyset, d) \nonumber \\
    & + \lambda\ \nabla_{\tilde{x}(t)}\ \mathcal{G}(\tilde{x}(t); t, y, d). \nonumber
\end{align}
The main focus of our work is designing a guidance energy $\mathcal{G}$ that biases the diffusion steps to produce a 3D-edited version of an input image. Section~\ref{sec:method} up to~\ref{sec:3d_edit} describes how we obtain the features needed for this energy, and Section~\ref{sec:generating_edited_image} defines the energy.

\paragraph{Latent Diffusion.}
In our experiments, we use a latent diffusion model where images $x_t$ contain features from a pre-trained latent space, rather than RGB values; 
but our method is also compatible with non-latent diffusion models, so we keep our description general.

\section{3D Edits for Diffusion Models}
\label{sec:method}

To perform 3D image edits that preserve both the realism and the identity of the original image, we apply the 3D edit to the feature spaces of intermediate layers in a pre-trained diffusion model, the \emph{activations} $\Psi$.
%
%
These activations describe the appearance and identity of objects in a generated image. We use activations in the decoder of the denoiser $\epsilon_\theta$, but only use layers with sufficient resolution to avoid inaccurate guidance; we use layers 2 and 3 (the last two layers) in the decoder of the StableDiffusion denoiser.
We denote activations of layer $i$ in denoising step $t$ as $\Psi_{i,t}$.

To apply a 3D edit to a given image $x$ with text prompt $y$, we proceed in three steps: (i)~we invert the image to reconstruct it with a diffusion model and save activations $\Psi$ of the generation process; (ii)~we apply the 3D edit to the activations; (iii)~we re-generate the image using the edited activations as guidance. In the remainder of this section we are going to describe these three steps in more detail.




 
\paragraph{Inverting the Input Image.}
Given an input image $x$, corresponding text prompt encoding $y$, we obtain the activations by inverting the image with our diffusion model using Null-Text Inversion~\cite{mokady2023null}. As our diffusion model is also conditioned by a depth map, the depth map can either be given as input (for example, from a synthetic 3D scene), or we estimate it using an existing monocular depth estimator~\cite{bhat2023zoedepth}. Note that the depth-conditioned diffusion model is tolerant to inaccuracies and noise in the depth map, as it was trained on estimated depth maps. The inversion gives us an initial noise $x_T$ and an updated null-text encoding $\emptyset'$ that we can use to reconstruct the input image $x$ in an inference pass $x_T, \dots, x_0$ of the diffusion model, such that $x_0 \approx x$. During inference pass, we record activations $\Psi_{i, t}$ of all relevant layers $i$ and time steps $t$.


\begin{figure}[t]
    \centering
    \includegraphics[width=\linewidth]{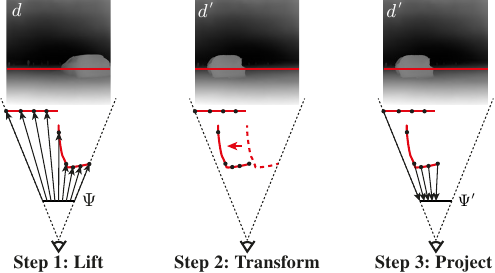}
    \caption{\textbf{3D edit of depth and activations.} (i)~Activations $\Psi$ are lifted to 3D surfaces; (ii)~the depth $d$ is 3D-transformed, together with the lifted $\Psi$; (iii)~the transformed $\Psi$ are projected back to the image plane. Section~\ref{sec:3d_edit} gives a detailed description of these steps.}
    \label{fig:3dedit}
\end{figure}

\subsection{Performing the 3D-Edit}
\label{sec:3d_edit}

To perform a 3D edit in our 2D image $x$, we define a simple warping mechanism $\mathcal{W}$. 
Given a flow field $F: [0, 1]^2 \rightarrow \mathbb{R}^2$, and a signal $X: [0,1]^2 \rightarrow C$,
$\mathcal{W}$ is defined as
\begin{equation}
    \mathcal{W}[X, F](u) = X(u - F(u)),
\end{equation}
where $u$ is a coordinate in $[0,1]^2$. This operator will be used throughout our method to warp signals defined on a 2D domain, such as attention maps and activations.

\paragraph{3D-aware Flow Field.}
A key differentiating factor of our approach
is that we compute the flow field $F$ in a 3D-aware manner.
This is done in three steps (see Figure~\ref{fig:3dedit}).

\emph{Step 1 - Lift:}
the lifting function $L_d:[0, 1]^2 \rightarrow \mathbb{R}^3$ assigns a 3D coordinate
to every point in the domain of the 2D signal $X$, based on the depth map $d$. Following~\cite{bhat2023zoedepth}, we assume a fixed horizontal field of view (fov) of $55$ degrees.

\emph{Step 2 - Transform:}
the user defines a function $T: \mathbb{R}^3 \rightarrow \mathbb{R}^3$ that modifies the position of the points in 3D space.
Groups of 3D points that correspond to an object can be identified with the help of image segmentation models. In our experiments, we opt for an open-set segmentation using SAM~\cite{kirillov2023segany}, starting from a bounding box found with Grounding DINO~\cite{liu2023grounding}. This gives the user the option to select the object of interest interactively, by describing the object with a text prompt and selecting one of the resulting candidate segments. We call the mask of the selected segment the \emph{object mask} $M_o$. Lifting $M_o$ to 3D identifies the 3D points corresponding to the object of interest, which can then be manipulated by the user with a rigid 3D transformation. All other 3D points remain unchanged.

\emph{Step 3 - Project:}
the projection function  $P: \mathbb{R}^3 \rightarrow \mathbb{R}^2$ projects 3D coordinates back to the 2D image plane assuming the same camera parameters as the lifting function.

By composing all three steps, we define an operation that transforms 2D coordinates in the original image to the corresponding position in the edited image. Our 3D-aware flow $F$ is based on the inverse of this operation:
\begin{equation}
F(u) = u - (P \circ T \circ L_d)^{-1} (u).
\end{equation}
Note that the inverse may not be defined for some 2D coordinates, since the operation $P \circ T \circ L_d$ is not always bijective; it may create overlapping regions (like occlusions) and holes (like disocclusions). We handle overlapping regions by picking the coordinates closest to the camera. To handle holes, we create a \emph{valid mask} $M_v$ of regions that are not holes $M_v \coloneqq \mathds{1}_{\text{range}(P \circ T \circ L_d)}$, and only guide regions inside this mask when generating the edited image.

\paragraph{Edited Maps.}
We warp the activations $\Psi_{i, t}$ with this 3D-aware flow field to get edited activations $\Psi_{i, t}'$:
\begin{equation}
    \Psi'_{i, t} \coloneqq \mathcal{W}[\rho(\Psi_{i, t}), F]
\end{equation}
where $\rho$ denotes bilinear interpolation.

\paragraph{Edited Depth.}
To obtain an edited depth map $d'$, we separately construct the edited depth for the transformed object $d'_o$ and for the remaining (static) scene $d'_b$, before re-compositing them. Treating them separately allows us to inpaint any holes in the static part of the scene that might be created by the 3D edit by leveraging the prior of a large 2D diffusion model.

Specifically, the depth for the static part of the scene $d'_b$ is obtained by removing the transformed object from the image $x$ using an existing object removal method~\cite{suvorov2021resolution}, with the object mask $M_o$ as input, resulting in an image $x_b$ without the transformed object. $d'_b$ is then estimated from $x_b$ using a monocular depth estimator~\cite{bhat2023zoedepth}.
%
%
%
We obtain depth of the transformed object $d'_o$ from the distance between the camera and the transformed 3D points $T \circ L_d$:
\begin{equation}
    d'_o(u) \coloneqq \|\mathcal{W}[T \circ L_d, F](u) \|_2,
\end{equation}
where we assume that the camera is at the origin.

The transformed object depth $d'_o$ and the depth of the remaining scene $d'_b$ are then composited seamlessly using Poisson Image Editing~\cite{perez2023poisson} to obtain the edited depth $d'$.

\begin{figure*}[t!]
    \centering
    \includegraphics[width=\textwidth]{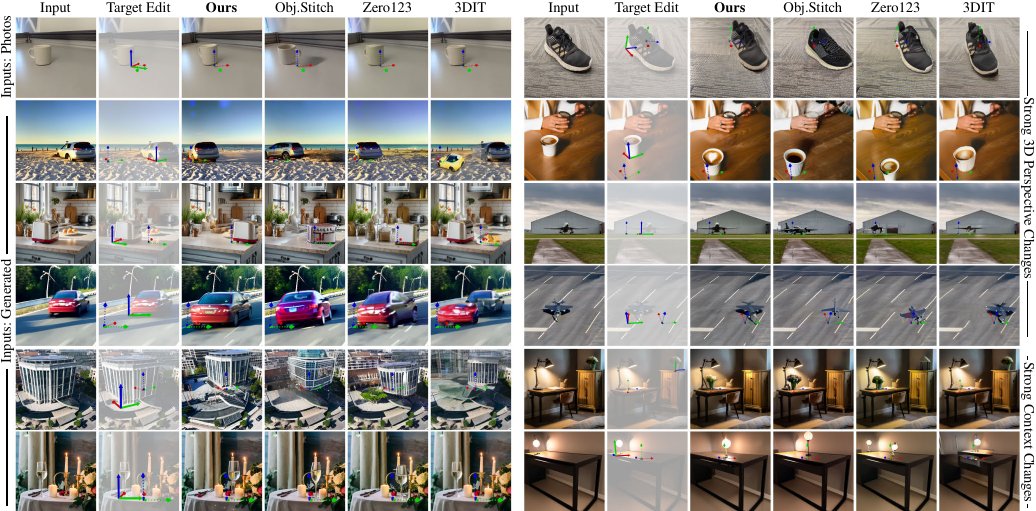}
    \caption{\textbf{Qualitative comparison.} We compare our method to three baselines. A target edit (from solid to dotted axes) is applied to an object from the input image . We show that our method better achieves our three goals: identity preservation, edit adherence, and plausibility.}
    \label{fig:main_results_qualtitative_comparison}
    \vspace{-5pt}
\end{figure*}

\subsection{Generating the Edited Image}
\label{sec:generating_edited_image}

We generate the edited image $x'_0$
using a diffusion process that is conditioned on the text prompt $y$ and the edited depth $d'$, and uses the initial noise $x_T$ and the null-text $\emptyset$ obtained from the inversion described at the start of this section.
Without any additional guidance, the resulting image would closely approximate the input image $x$. Thus, we guide the diffusion process to follow the edited activations $\Psi'$.
Guiding the diffusion process to follow the edited activations encourages the resulting image to preserve the identity of objects from the original scene and to follow the edited object layout.
We add two energy terms.

The \emph{object guidance} energy $\mathcal{G}_o$ focuses on the edited object only. It is the L2 distance between the activations of the diffusion process $\Psi^e$ and the edited activations $\Psi'$:
\begin{equation}
\label{eq:foreground_guidance}
    \mathcal{G}_o \coloneqq \sum_{i,t} w^o_{i,t} \sum_u \big(M'_o (\Psi^e_{i,t} - \Psi'_{i,t})\big)^2(u),
\end{equation}
%
where $\Psi^e$ are the activations of the denoiser in the diffusion process, and $M'_o \coloneqq \mathcal{W}[M_o, F] \cdot M_v$ is the valid part of the warped object mask, i.e. the mask of the foreground object in the edited image.
$w^o_{i,t}$ is a per-step and per-layer weight we set according to a schedule. See below for details.


The \emph{background guidance} energy $\mathcal{G}_b$ is defined similarly to the object guidance energy, but focuses on the static part of the scene only:
\begin{equation}
    \mathcal{G}_b \coloneqq \sum_{i,t} w^b_{i,t}\ \frac{\big(\sum_{u}M'_b \Psi^e_{i,t}(u) - \sum_{u}M'_b\Psi'_{i,t}(u)\big)^2}{\sum_u M'_b(u)},
\end{equation}
%
where $M'_b \coloneqq 1 - M'_o$ and $w^b_t$ is set according to a similar schedule as $w^o_{i, t}$, see below for details. We compare the average of the activations over the image, as we expect some parts of the static scene to change, for example due lighting or shadows, disocclusions, etc.

We set weights $w^o_{i, t}$ and $w^b_{i, t}$ according to a \emph{guidance schedule}.
(i)~We guide only up until time step $38/50$, and then zero the guidance.
%
(ii)~We cycle between guiding different layers in each time step, guiding layer 3 in the first step, layer 2 in the second step, both layers in the third step, and repeat this cycle in subsequent steps.
%
(iii)~Finally, we adjust the relative weighting of $w^o_{i, t}$ and $w^b_{i, t}$ according to the desired level of foreground and background preservation.
%
Section~\ref{sec:results} discusses and motivates these design choices and gives examples of different settings in an ablation.


We define the final guidance energy $\mathcal{G}$ as, 
    $\mathcal{G} \coloneqq \mathcal{G}_o + \mathcal{G}_b$.
 We use the gradient of this guidance energy to bias each step of the diffusion process, as described in Eq.\eqref{eq:guidance} as, 
   $ \epsilon_\theta^\mathcal{G}(x_t; t, y, \emptyset', d)$
 resulting in the edited image $x'_0$.

\section{Results}
\label{sec:results}



\paragraph{Datasets.}
We created two datasets to evaluate our method. The \texttt{PhotoGen} dataset consists of $31$ edits in $26$ images images (five of the images have two different edits) that we either generated, photographed, or licensed and downloaded. It contains $5$ photographs and $21$ generated images. In the \texttt{Benchmark} dataset we aim at minimizing our (the author's) bias in the choice of images and edits. It consists of $50$ edits in $50$ images (one edit per image) that we generated with a depth-to-image diffusion model using depth from synthetic 3D scenes. The 3D scenes are generated automatically by randomly choosing a 3D asset from 10 categories in the ModelNet40~\cite{wu20153d} dataset, and placing the asset at a random location on a ground plane. Edits are randomly chosen from 3D translations and 3D rotations. Both the parameters of the initial placement and of the edit are constrained to ensures objects remain withing the view frustum and exhibit a limited amount of disocclusion after the edit.

\paragraph{Baselines.}
We compare to several state-of-the-art methods that share our goal of image editing with generative models. \emph{Object3DIT}~\cite{michel2023object} fintunes Zero123~\cite{zero123} with synthetic data to enable either 3D rotations, scaling, or translation on a ground plane. \emph{ObjectStitch}~\cite{song2023objectstitch} allows transplanting objects from one image to a given 2D position in another image; we use it to transplant objects to a different location in the same image and remove the original, unedited object using the same object removal method~\cite{suvorov2021resolution} we use in our depth edit. We create another baseline that uses \emph{Zero123} to get a novel view of the foreground object, removes the original foreground object~\cite{suvorov2021resolution}, moves the novel view to a new image location, and inpaints a 15-pixel-wide region around the novel view using Firefly~\cite{FireFly} to improve image coherence. We also experimented with Diffusion Self-Guidance~\cite{epstein2023diffusion}, but found that the public code performs far worse in terms of identity preservation and edit adherence than the published version (which uses the proprietary Imagen~\cite{saharia2022photorealistic}), as confirmed by the authors. For fairness, we instead show an ablation that comes close to this method. 

\paragraph{Qualitative Comparison.}
%
See Figure~\ref{fig:main_results_qualtitative_comparison}. ObjectStitch generates scenes with good plausibility, but as it does not provide 3D controls, we observe low edit adherence (i.e., the output does not match the target edit); we also observe relatively low identity preservation. Zero123 and 3DIT have better identity preservation, but Zero123 struggles to generate good novel views for objects that are for from the dataset it was finetuned on (like the somewhat blurry red car), and the fixed inpainting region limits the plausibility of secondary effects like shadows and lighting in the edited scene (see for example the lack of shadows for the coffee cup). 3DIT is biased even more strongly than Zero123 by the synthetic scenes it was fintuned on, in which objects have a limited range of sizes and types, and are viewed from a limited range of angles. 
3DIT lacks generalization, for example, it fails to perform an edit (e.g. the wine glasses), or places a novel object into the scene (e.g. the car on the beach). Our method shows better identity preservation and edit adherence due to our detailed 3D-aware guidance, and better plausibility since we do not perform any fine-tuning that would bias the prior of the pre-trained diffusion model towards more constrained scene types.
%
The supplementary provides comparisons on the full \texttt{PhotoGen} dataset.
%
Figure~\ref{fig:stop_motion} shows multiple intermediate edits along two edit trajectories, demonstrating the consistency and identity preservation of our method.


\paragraph{User Study.}
We quantitatively compare our method to all baselines in terms of the three desirable goals: identity preservation, edit adherence, and plausibility, with a user study on a subset of 11 edits in 11 images from our \texttt{PhotoGen} dataset, including one photograph. 
%
We separately evaluate each desirable goal by showing users pairs of images and asking them to select the image that better fulfills the goal.
We form 66 random image pairs, where each pair compares a random result from our method to the corresponding result from a random baseline. We split these 66 pair into 3 groups of 22 pairs each, one group for each goal. A total of 22 users participated in the study, with mixed expertise in image editing. Each user compared all 22 pairs for each goal (484 data points per goal, and an average of 161 data points for each of the three method pairings). For the plausibility goal, we additionally compare to the original input image as an upper bound for the achievable plausibility (average of 121 data points per method pair). Both the order of
pairs for each goal, and
of methods in each pair was randomized. See supplemental for details.

Results are shown in Figure~\ref{fig:user_study}. We can see that users clearly preferred our method over the baselines in all three goals. In some cases, users even found our results to be more plausible than the original input images, although, as we would expect, the original images were still more plausible on average. The results support our observations from the qualitative results: ObjectStitch has good plausibiliy, but relatively low identity preservation and edit adherence. 3DIT and Zero123 have better identity preservation and edit adherence, but lower plausibility. 3DIT has especially low plausibility due to its biased diffusion prior.

\begin{figure}[t!]
    \centering
    \includegraphics[width=\columnwidth]{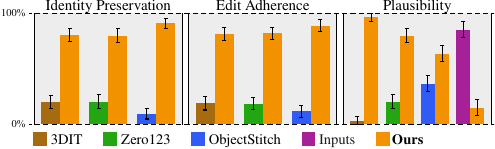}
    \caption{\textbf{User study.} We compare how well each method achieves our three main goals.
    Each pair of bars show the percentage of users that preferred our method (orange) or a baseline (other color) with 95$\%$ confidence intervals. The \emph{inputs} bar represents an upper bound to the plausibility of an edited image.}
    \label{fig:user_study}
\end{figure}

\begin{figure}[b!]
    \centering
    \includegraphics[width=\columnwidth]{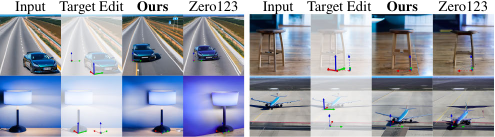}
    \caption{\textbf{Synthetic benchmark.} Comparison to Zero123~\cite{zero123} on a few examples of our \texttt{Benchmark} dataset.}
    \label{fig:benchmark}
\end{figure}

\paragraph{Synthetic Benchmark.}
%
We automatically generate images and edits to reduce selection bias in the results. We use synthetic depth, which comes as a by-product of the automatic image generation. This ensures that scene depth is reasonable, which factors out the influence of errors in the depth estimate from our experiments. The full benchmark on all 50 scenes is given in the supplementary, Figure~\ref{fig:benchmark} shows a qualitative comparison on four scenes. We see good identity preservation, edit adherence and plausibility in these scenes, suggesting that our method works robustly on random scenes given a reasonable depth.
We additionally provide a quantitative comparison to Zero123 (the only comparable baseline that supports both 3D translation and rotation) on 10 of the scenes that evaluates edit adherence and identity preservation. \emph{Edit adherence} is evaluated using the Intersection over Union (IoU) between the a SAM-based~\cite{kirillov2023segany} segmentation of the foreground object in the edited image and a ground truth segmentation mask obtained from the synthetic depth.
Our method consistently outperforms Zero123 with an IoU of 0.87 compared to Zero123's 0.59. \emph{Identity Preservation} is evaluated using a cycle consistency metric that transform the edited image back to the original object configuration and measures the difference to the original input image using both the L1 distance and LPIPS~\cite{zhang2018unreasonable}. Our method achieves a superior L1$\downarrow$/LPIPS$\downarrow$ of 0.084/0.22 compared to Zero123's 0.098/0.30. As many pixels are left untouched by the edit, this is a significant improvement.

\begin{figure}[t!]
    \centering
    \includegraphics[width=\columnwidth]{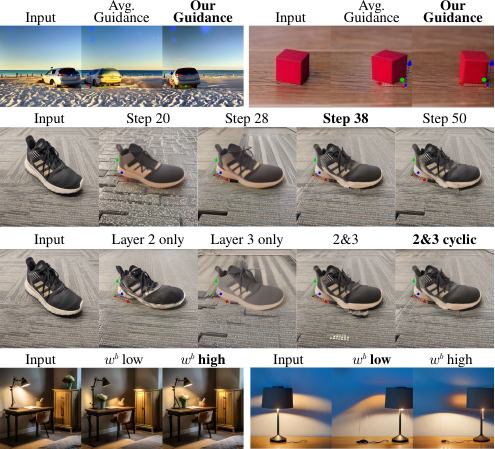}
    \caption{\textbf{Ablation Study.} We show the effect of several design choices of our method. See the \emph{Ablation Study} paragraph for details.}
    \label{fig:ablation}
\end{figure}
\paragraph{Ablation Study.}

We ablate four design choices, see Figure~\ref{fig:ablation}.
(i)~In the first row, we show the effect of using our form of local guidance for the foreground vs. using the average over the foreground region in Eq.~\ref{eq:foreground_guidance}, similar to Diffusion Self-Guidance~\cite{epstein2023diffusion}. We can see that our local guidance significantly improves identity preservation and edit adherence.
(ii)~In the second row, we guide up to a different maximum number of steps. Intuitively, the last time steps allow the model to reconcile the edited object with the scene, by creating details such as contact shadows and lighting. We found that giving the model space to do this reconciliation without guidance increases plausibility. Guiding too few steps, on the other hand, reduces identity preservation.
(iii)~In the third row, we show the effect of using different choices of layers in the guidance schedule for each time step. Guiding the second layer of the denoiser decoder only tends to preserve texture style, but loses some identity preservation and edit adherence, while guiding the third layer only tends to have the opposite effect.
Ideally we want to preserve all three properties, but we found that guiding both layers introduces artifacts, possibly because the guidance of different layers can be contradictory to some extent. Our cyclic schedule allows us to guide both layers with reduced artifacts.
%
(iv)~The last row examines the balance of foreground and background weights $w^o$ and $w^b$. In both scenes, the lighting of the foreground object and background are at odds (e.g. on the left, the vase is originally unlit and the background at the target position has strong lighting). Setting $w^b$ low relative to $w^o$ preserves foreground lighting, but changes background lighting, and vice-versa for high $w^b$.











\begin{figure}[b!]
    \centering
    \includegraphics[width=\columnwidth]{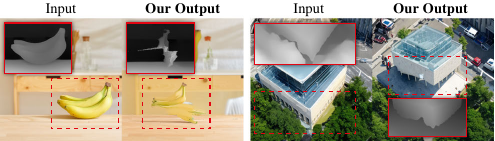}
    \caption{\textbf{Limitations.} Our method relies on a reasonable depth map. Large edits that reveal strong distortions of a depth estimate or missing parts of the depth result in low-quality output.} 
    \label{fig:limitations}
\end{figure}

\section{Conclusion}

We have presented \name to enable 
3D-aware object level edits
on 2D images, which may be generated or real photographs. We do not require additional training or 3D supervision data, and avoid explicitly solving the inverse graphics problem. We demonstrated that by lifting intermediate diffusion activations to 3D using estimated depth, and transforming the activations with user-specified 3D edits, one can produce realistic images with a good balance between plausibility and identity control while respecting the target edits. In our extensive tests, we demonstrated the superiority of our proposed approach against other contemporary baselines using both quantitative and qualitative metrics. We believe that \name fills an important gap in current generative image workflows. 

\begin{figure}[t]
    \centering
    \includegraphics[width=\columnwidth]{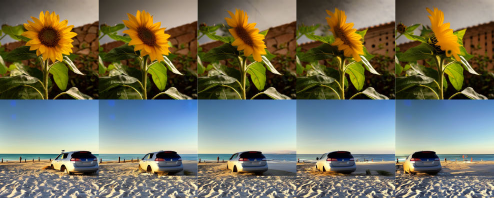}
    \caption{\textbf{Stop-motion edits.} Intermediate edits along two edit trajectories demonstrate consistency and identity preservation.}
    \label{fig:stop_motion}
\end{figure}

\paragraph{Limitations and Future Work.}
Although our method is robust to the quality of the estimated depths, which are often warped strongly in view direction, large edits that make this warping apparent, and edits that reveal parts of the objects hidden in the original view may give undesirable results (see Figure~\ref{fig:limitations}).
In the future, we would like to regularize the problem using shape priors to infill the estimated depth maps in occluded regions. One exciting option would be to use recent image-to-Nerf models~\cite{mi2022im2nerf,hong2023lrm} to perform such a regularization. 
Another limitation of our method is that identity preservation, while better than existing methods, is still not perfect.
In the future, we expect generative image models to also produce additional channels (e.g., albedo, normal, specular, illumination) that would allow more physically grounded control over object identity that is hard to achieve directly using only RGB information. 
Finally, we would like to extend our method to produce video output by animating 3D edits similar to Figure~\ref{fig:stop_motion}, but with more frames. The challenging part will be ensuring temporal smoothness while preserving object identity without additional training. We expect to use pre-trained video diffusion models~\cite{ho2022video}. 

{
    \small
    \bibliographystyle{ieeenat_fullname}
    \bibliography{main}

\begin{thebibliography}{61}
\providecommand{\natexlab}[1]{#1}
\providecommand{\url}[1]{\texttt{#1}}
\expandafter\ifx\csname urlstyle\endcsname\relax
  \providecommand{\doi}[1]{doi: #1}\else
  \providecommand{\doi}{doi: \begingroup \urlstyle{rm}\Url}\fi

\bibitem[{Adobe}()]{FireFly}
{Adobe}.
\newblock Firefly.

\bibitem[Avrahami et~al.(2022)Avrahami, Lischinski, and Fried]{avrahami2022blended}
Omri Avrahami, Dani Lischinski, and Ohad Fried.
\newblock Blended diffusion for text-driven editing of natural images.
\newblock In \emph{Proceedings of the IEEE/CVF Conference on Computer Vision and Pattern Recognition}, pages 18208--18218, 2022.

\bibitem[Avrahami et~al.(2023)Avrahami, Hayes, Gafni, Gupta, Taigman, Parikh, Lischinski, Fried, and Yin]{spatext}
Omri Avrahami, Thomas Hayes, Oran Gafni, Sonal Gupta, Yaniv Taigman, Devi Parikh, Dani Lischinski, Ohad Fried, and Xi Yin.
\newblock Spatext: Spatio-textual representation for controllable image generation.
\newblock In \emph{CVPR}, 2023.

\bibitem[Bar-Tal et~al.(2023)Bar-Tal, Yariv, Lipman, and Dekel]{multidiffusion}
Omer Bar-Tal, Lior Yariv, Yaron Lipman, and Tali Dekel.
\newblock Multidiffusion: Fusing diffusion paths for controlled image generation.
\newblock \emph{Int. Conf. Machine Learning}, 2023.

\bibitem[Betker et~al.()Betker, Goh, Jing, Brooks, Wang, Li, Ouyang, Zhuang, Lee, Guo, Manassra, Dhariwal, Chu, Jiao, and Ramesh]{dalle3}
James Betker, Gabriel Goh, Li Jing, Tim Brooks, Jianfeng Wang, Linjie Li, Long Ouyang, Juntang Zhuang, Joyce Lee, Yufei Guo, Wesam Manassra, Prafulla Dhariwal, Casey Chu, Yunxin Jiao, and Aditya Ramesh.
\newblock Improving image generation with better captions.

\bibitem[Bhat et~al.(2023)Bhat, Birkl, Wofk, Wonka, and M{\"u}ller]{bhat2023zoedepth}
Shariq~Farooq Bhat, Reiner Birkl, Diana Wofk, Peter Wonka, and Matthias M{\"u}ller.
\newblock Zoedepth: Zero-shot transfer by combining relative and metric depth.
\newblock \emph{arXiv preprint arXiv:2302.12288}, 2023.

\bibitem[Brooks et~al.(2023)Brooks, Holynski, and Efros]{instructpix2pix}
Tim Brooks, Aleksander Holynski, and Alexei~A. Efros.
\newblock Instructpix2pix: Learning to follow image editing instructions.
\newblock In \emph{CVPR}, 2023.

\bibitem[Chang et~al.(2023)Chang, Zhang, Barber, Maschinot, Lezama, Jiang, Yang, Murphy, Freeman, Rubinstein, et~al.]{chang2023muse}
Huiwen Chang, Han Zhang, Jarred Barber, AJ Maschinot, Jose Lezama, Lu Jiang, Ming-Hsuan Yang, Kevin Murphy, William~T Freeman, Michael Rubinstein, et~al.
\newblock Muse: Text-to-image generation via masked generative transformers.
\newblock \emph{Int. Conf. Machine Learning}, 2023.

\bibitem[Chen et~al.(2023)Chen, Laina, and Vedaldi]{chen2023training}
Minghao Chen, Iro Laina, and Andrea Vedaldi.
\newblock Training-free layout control with cross-attention guidance.
\newblock \emph{arXiv preprint arXiv:2304.03373}, 2023.

\bibitem[Chen et~al.(2009)Chen, Cheng, Tan, Shamir, and Hu]{sketch2photo}
Tao Chen, Ming-Ming Cheng, Ping Tan, Ariel Shamir, and Shimin Hu.
\newblock Sketch2photo: internet image montage.
\newblock \emph{ACM Transactions on Computer Graphics}, 2009.

\bibitem[Crowson et~al.(2022)Crowson, Biderman, Kornis, Stander, Hallahan, Castricato, and Raff]{vqganclip}
Katherine Crowson, Stella~Rose Biderman, Daniel Kornis, Dashiell Stander, Eric Hallahan, Louis Castricato, and Edward Raff.
\newblock Vqgan-clip: Open domain image generation and editing with natural language guidance.
\newblock In \emph{ECCV}, 2022.

\bibitem[Ding et~al.(2021)Ding, Yang, Hong, Zheng, Zhou, Yin, Lin, Zou, Shao, Yang, and Tang]{cogview}
Ming Ding, Zhuoyi Yang, Wenyi Hong, Wendi Zheng, Chang Zhou, Da Yin, Junyang Lin, Xu Zou, Zhou Shao, Hongxia Yang, and Jie Tang.
\newblock Cogview: Mastering text-to-image generation via transformers.
\newblock In \emph{NeurIPS}, 2021.

\bibitem[Epstein et~al.(2023)Epstein, Jabri, Poole, Efros, and Holynski]{epstein2023diffusion}
Dave Epstein, Allan Jabri, Ben Poole, Alexei~A Efros, and Aleksander Holynski.
\newblock Diffusion self-guidance for controllable image generation.
\newblock \emph{arXiv preprint arXiv:2306.00986}, 2023.

\bibitem[Frans et~al.(2022)Frans, Soros, and Witkowski]{clipdraw}
Kevin Frans, Lisa Soros, and Olaf Witkowski.
\newblock Clipdraw: Exploring text-to-drawing synthesis through language-image encoders.
\newblock In \emph{NeurIPS}, 2022.

\bibitem[Gafni et~al.(2022)Gafni, Polyak, Ashual, Sheynin, Parikh, and Taigman]{makeascene}
Oran Gafni, Adam Polyak, Oron Ashual, Shelly Sheynin, Devi Parikh, and Yaniv Taigman.
\newblock Make-a-scene: Scene-based text-to-image generation with human priors.
\newblock 2022.

\bibitem[Ho and Salimans(2022)]{ho2022classifier}
Jonathan Ho and Tim Salimans.
\newblock Classifier-free diffusion guidance.
\newblock \emph{arXiv preprint arXiv:2207.12598}, 2022.

\bibitem[Ho et~al.(2020)Ho, Jain, and Abbeel]{ho2020denoising}
Jonathan Ho, Ajay Jain, and Pieter Abbeel.
\newblock Denoising diffusion probabilistic models.
\newblock \emph{Advances in neural information processing systems}, 33:\penalty0 6840--6851, 2020.

\bibitem[Ho et~al.(2022)Ho, Salimans, Gritsenko, Chan, Norouzi, and Fleet]{ho2022video}
Jonathan Ho, Tim Salimans, Alexey Gritsenko, William Chan, Mohammad Norouzi, and David~J Fleet.
\newblock Video diffusion models.
\newblock \emph{arXiv:2204.03458}, 2022.

\bibitem[Hoiem et~al.(2005)Hoiem, Efros, and Hebert]{photopopup}
Derek Hoiem, Alexei~A. Efros, and Martial Hebert.
\newblock Automatic photo pop-up.
\newblock In \emph{ACM Transactions on Computer Graphics}, 2005.

\bibitem[Hong et~al.(2023)Hong, Zhang, Gu, Bi, Zhou, Liu, Liu, Sunkavalli, Bui, and Tan]{hong2023lrm}
Yicong Hong, Kai Zhang, Jiuxiang Gu, Sai Bi, Yang Zhou, Difan Liu, Feng Liu, Kalyan Sunkavalli, Trung Bui, and Hao Tan.
\newblock Lrm: Large reconstruction model for single image to 3d, 2023.

\bibitem[Jampani et~al.(2021)Jampani, Chang, Sargent, Kar, Tucker, Krainin, Kaeser, Freeman, Salesin, Curless, and Liu]{slide}
Varun Jampani, Huiwen Chang, Kyle Sargent, Abhishek Kar, Richard Tucker, Michael Krainin, Dominik Kaeser, William~T Freeman, David Salesin, Brian Curless, and Ce Liu.
\newblock Slide: Single image 3d photography with soft layering and depth-aware inpainting.
\newblock In \emph{CVPR}, 2021.

\bibitem[Kholgade et~al.(2014{\natexlab{a}})Kholgade, Simon, Efros, and Sheikh]{kholgade20143d}
Natasha Kholgade, Tomas Simon, Alexei Efros, and Yaser Sheikh.
\newblock 3d object manipulation in a single photograph using stock 3d models.
\newblock \emph{ACM Transactions on graphics (TOG)}, 33\penalty0 (4):\penalty0 1--12, 2014{\natexlab{a}}.

\bibitem[Kholgade et~al.(2014{\natexlab{b}})Kholgade, Simon, Efros, and Sheikh]{objmanipulation}
Natasha Kholgade, Tomas Simon, Alexei~A. Efros, and Yaser Sheikh.
\newblock 3d object manipulation in a single photograph using stock 3d models.
\newblock \emph{ACM Transactions on Computer Graphics}, 2014{\natexlab{b}}.

\bibitem[Kirillov et~al.(2023)Kirillov, Mintun, Ravi, Mao, Rolland, Gustafson, Xiao, Whitehead, Berg, Lo, Doll{\'a}r, and Girshick]{kirillov2023segany}
Alexander Kirillov, Eric Mintun, Nikhila Ravi, Hanzi Mao, Chloe Rolland, Laura Gustafson, Tete Xiao, Spencer Whitehead, Alexander~C. Berg, Wan-Yen Lo, Piotr Doll{\'a}r, and Ross Girshick.
\newblock Segment anything.
\newblock \emph{arXiv:2304.02643}, 2023.

\bibitem[Li et~al.(2023{\natexlab{a}})Li, Wang, Wang, and Xiong]{li2023image}
Jiajie Li, Jian Wang, Chen Wang, and Jinjun Xiong.
\newblock Image harmonization with diffusion model, 2023{\natexlab{a}}.

\bibitem[Li et~al.(2023{\natexlab{b}})Li, Liu, Wu, Mu, Yang, Gao, Li, and Lee]{li2023gligen}
Yuheng Li, Haotian Liu, Qingyang Wu, Fangzhou Mu, Jianwei Yang, Jianfeng Gao, Chunyuan Li, and Yong~Jae Lee.
\newblock Gligen: Open-set grounded text-to-image generation.
\newblock In \emph{CVPR}, pages 22511--22521, 2023{\natexlab{b}}.

\bibitem[Liu et~al.(2023{\natexlab{a}})Liu, Wu, Van~Hoorick, Tokmakov, Zakharov, and Vondrick]{zero123}
Ruoshi Liu, Rundi Wu, Basile Van~Hoorick, Pavel Tokmakov, Sergey Zakharov, and Carl Vondrick.
\newblock Zero-1-to-3: Zero-shot one image to 3d object.
\newblock In \emph{ICCV}, 2023{\natexlab{a}}.

\bibitem[Liu et~al.(2023{\natexlab{b}})Liu, Zeng, Ren, Li, Zhang, Yang, Li, Yang, Su, Zhu, et~al.]{liu2023grounding}
Shilong Liu, Zhaoyang Zeng, Tianhe Ren, Feng Li, Hao Zhang, Jie Yang, Chunyuan Li, Jianwei Yang, Hang Su, Jun Zhu, et~al.
\newblock Grounding dino: Marrying dino with grounded pre-training for open-set object detection.
\newblock \emph{arXiv preprint arXiv:2303.05499}, 2023{\natexlab{b}}.

\bibitem[Lugmayr et~al.(2022)Lugmayr, Danelljan, Romero, Yu, Timofte, and Gool]{lugmayr2022repaint}
Andreas Lugmayr, Martin Danelljan, Andres Romero, Fisher Yu, Radu Timofte, and Luc~Van Gool.
\newblock Repaint: Inpainting using denoising diffusion probabilistic models, 2022.

\bibitem[Mi et~al.(2022)Mi, Kundu, Ross, Dellaert, Snavely, and Fathi]{mi2022im2nerf}
Lu Mi, Abhijit Kundu, David Ross, Frank Dellaert, Noah Snavely, and Alireza Fathi.
\newblock im2nerf: Image to neural radiance field in the wild, 2022.

\bibitem[Michel et~al.(2023)Michel, Bhattad, VanderBilt, Krishna, Kembhavi, and Gupta]{michel2023object}
Oscar Michel, Anand Bhattad, Eli VanderBilt, Ranjay Krishna, Aniruddha Kembhavi, and Tanmay Gupta.
\newblock Object 3dit: Language-guided 3d-aware image editing.
\newblock \emph{arXiv preprint arXiv:2307.11073}, 2023.

\bibitem[Mokady et~al.(2023)Mokady, Hertz, Aberman, Pritch, and Cohen-Or]{mokady2023null}
Ron Mokady, Amir Hertz, Kfir Aberman, Yael Pritch, and Daniel Cohen-Or.
\newblock Null-text inversion for editing real images using guided diffusion models.
\newblock In \emph{Proceedings of the IEEE/CVF Conference on Computer Vision and Pattern Recognition}, pages 6038--6047, 2023.

\bibitem[Mori et~al.(1999)Mori, Takahashi, and Oka]{mori1999}
Y. Mori, H. Takahashi, and R. Oka.
\newblock Image-to-word transformation based on dividing and vector quantizing images with words.
\newblock In \emph{MISRM'99 First International Workshop on Multimedia Intelligent Storage and Retrieval Management}, 1999.

\bibitem[Mou et~al.(2023{\natexlab{a}})Mou, Wang, Song, Shan, and Zhang]{mou2023dragondiffusion}
Chong Mou, Xintao Wang, Jiechong Song, Ying Shan, and Jian Zhang.
\newblock Dragondiffusion: Enabling drag-style manipulation on diffusion models.
\newblock \emph{arXiv preprint arXiv:2307.02421}, 2023{\natexlab{a}}.

\bibitem[Mou et~al.(2023{\natexlab{b}})Mou, Wang, Xie, Wu, Zhang, Qi, Shan, and Qie]{t2i}
Chong Mou, Xintao Wang, Liangbin Xie, Yanze Wu, Jian Zhang, Zhongang Qi, Ying Shan, and Xiaohu Qie.
\newblock T2i-adapter: Learning adapters to dig out more controllable ability for text-to-image diffusion models.
\newblock \emph{arXiv preprint arXiv:2302.08453}, 2023{\natexlab{b}}.

\bibitem[Nguyen-Phuoc et~al.(2020)Nguyen-Phuoc, Richardt, Mai, Yang, and Mitra]{BlockGAN2020}
Thu Nguyen-Phuoc, Christian Richardt, Long Mai, Yong-Liang Yang, and Niloy Mitra.
\newblock Blockgan: Learning 3d object-aware scene representations from unlabelled images.
\newblock In \emph{NeurIPS}, 2020.

\bibitem[Nichol et~al.(2021)Nichol, Dhariwal, Ramesh, Shyam, Mishkin, McGrew, Sutskever, and Chen]{nichol2021glide}
Alex Nichol, Prafulla Dhariwal, Aditya Ramesh, Pranav Shyam, Pamela Mishkin, Bob McGrew, Ilya Sutskever, and Mark Chen.
\newblock Glide: Towards photorealistic image generation and editing with text-guided diffusion models.
\newblock \emph{arXiv preprint arXiv:2112.10741}, 2021.

\bibitem[Niklaus et~al.(2019)Niklaus, Mai, Yang, and Liu]{kenburns}
Simon Niklaus, Long Mai, Jimei Yang, and Feng Liu.
\newblock 3d ken burns effect from a single image.
\newblock \emph{ACM Transactions on Computer Graphics}, 2019.

\bibitem[Pan et~al.(2023)Pan, Tewari, Leimk{\"u}hler, Liu, Meka, and Theobalt]{pan2023drag}
Xingang Pan, Ayush Tewari, Thomas Leimk{\"u}hler, Lingjie Liu, Abhimitra Meka, and Christian Theobalt.
\newblock Drag your gan: Interactive point-based manipulation on the generative image manifold.
\newblock In \emph{ACM SIGGRAPH 2023 Conference Proceedings}, pages 1--11, 2023.

\bibitem[Park et~al.(2019)Park, Liu, Wang, and Zhu]{park2019semantic}
Taesung Park, Ming-Yu Liu, Ting-Chun Wang, and Jun-Yan Zhu.
\newblock Semantic image synthesis with spatially-adaptive normalization.
\newblock In \emph{CVPR}, 2019.

\bibitem[Patashnik et~al.(2021)Patashnik, Wu, Shechtman, Cohen-Or, and Lischinski]{styleclip}
Or Patashnik, Zongze Wu, Eli Shechtman, Daniel Cohen-Or, and Dani Lischinski.
\newblock Styleclip: Text-driven manipulation of stylegan imagery.
\newblock In \emph{ICCV}, 2021.

\bibitem[P{\'e}rez et~al.(2023)P{\'e}rez, Gangnet, and Blake]{perez2023poisson}
Patrick P{\'e}rez, Michel Gangnet, and Andrew Blake.
\newblock Poisson image editing.
\newblock In \emph{Seminal Graphics Papers: Pushing the Boundaries, Volume 2}, pages 577--582. 2023.

\bibitem[Po and Wetzstein(2023)]{po2023compositional}
Ryan Po and Gordon Wetzstein.
\newblock Compositional 3d scene generation using locally conditioned diffusion.
\newblock \emph{arXiv preprint arXiv:2303.12218}, 2023.

\bibitem[Quattoni et~al.(2007)Quattoni, Collins, and Darrell]{quattoni2007learning}
Ariadna Quattoni, Michael Collins, and Trevor Darrell.
\newblock Learning visual representations using images with captions.
\newblock In \emph{CVPR}, 2007.

\bibitem[Radford et~al.(2021)Radford, Kim, Hallacy, Ramesh, Goh, Agarwal, Sastry, Askell, Mishkin, Clark, Krueger, and Sutskever]{clip}
Alec Radford, Jong~Wook Kim, Chris Hallacy, Aditya Ramesh, Gabriel Goh, Sandhini Agarwal, Girish Sastry, Amanda Askell, Pamela Mishkin, Jack Clark, Gretchen Krueger, and Ilya Sutskever.
\newblock Learning transferable visual models from natural language supervision.
\newblock In \emph{Int. Conf. Machine Learning}, 2021.

\bibitem[Ramesh et~al.(2022)Ramesh, Dhariwal, Nichol, Chu, and Chen]{ramesh2022dalle}
Aditya Ramesh, Prafulla Dhariwal, Alex Nichol, Casey Chu, and Mark Chen.
\newblock Hierarchical text-conditional image generation with clip latents.
\newblock \emph{arXiv preprint arXiv:2204.06125}, 2022.

\bibitem[Rombach et~al.(2022)Rombach, Blattmann, Lorenz, Esser, and Ommer]{rombach2022latentdiffusion}
Robin Rombach, Andreas Blattmann, Dominik Lorenz, Patrick Esser, and Bj{\"o}rn Ommer.
\newblock High-resolution image synthesis with latent diffusion models.
\newblock In \emph{Proceedings of the IEEE/CVF conference on computer vision and pattern recognition}, pages 10684--10695, 2022.

\bibitem[Saharia et~al.(2022{\natexlab{a}})Saharia, Chan, Chang, Lee, Ho, Salimans, Fleet, and Norouzi]{saharia2022palette}
Chitwan Saharia, William Chan, Huiwen Chang, Chris Lee, Jonathan Ho, Tim Salimans, David Fleet, and Mohammad Norouzi.
\newblock Palette: Image-to-image diffusion models.
\newblock In \emph{ACM SIGGRAPH 2022 Conference Proceedings}, pages 1--10, 2022{\natexlab{a}}.

\bibitem[Saharia et~al.(2022{\natexlab{b}})Saharia, Chan, Saxena, Li, Whang, Denton, Ghasemipour, Gontijo~Lopes, Karagol~Ayan, Salimans, et~al.]{saharia2022photorealistic}
Chitwan Saharia, William Chan, Saurabh Saxena, Lala Li, Jay Whang, Emily~L Denton, Kamyar Ghasemipour, Raphael Gontijo~Lopes, Burcu Karagol~Ayan, Tim Salimans, et~al.
\newblock Photorealistic text-to-image diffusion models with deep language understanding.
\newblock \emph{Advances in Neural Information Processing Systems}, 35:\penalty0 36479--36494, 2022{\natexlab{b}}.

\bibitem[Shih et~al.(2020)Shih, Su, Kopf, and Huang]{3dphotography}
Meng-Li Shih, Shih-Yang Su, Johannes Kopf, and Jia-Bin Huang.
\newblock 3d photography using context-aware layered depth inpainting.
\newblock In \emph{CVPR}, 2020.

\bibitem[Song et~al.(2020)Song, Meng, and Ermon]{song2020denoising}
Jiaming Song, Chenlin Meng, and Stefano Ermon.
\newblock Denoising diffusion implicit models.
\newblock \emph{arXiv preprint arXiv:2010.02502}, 2020.

\bibitem[Song et~al.(2023)Song, Zhang, Lin, Cohen, Price, Zhang, Kim, and Aliaga]{song2023objectstitch}
Yizhi Song, Zhifei Zhang, Zhe Lin, Scott Cohen, Brian Price, Jianming Zhang, Soo~Ye Kim, and Daniel Aliaga.
\newblock Objectstitch: Object compositing with diffusion model.
\newblock In \emph{Proceedings of the IEEE/CVF Conference on Computer Vision and Pattern Recognition}, pages 18310--18319, 2023.

\bibitem[Srivastava and Salakhutdinov(2012)]{srivastava2012}
Nitish Srivastava and Russ~R Salakhutdinov.
\newblock Multimodal learning with deep boltzmann machines.
\newblock In \emph{NeurIPS}, 2012.

\bibitem[Suvorov et~al.(2021)Suvorov, Logacheva, Mashikhin, Remizova, Ashukha, Silvestrov, Kong, Goka, Park, and Lempitsky]{suvorov2021resolution}
Roman Suvorov, Elizaveta Logacheva, Anton Mashikhin, Anastasia Remizova, Arsenii Ashukha, Aleksei Silvestrov, Naejin Kong, Harshith Goka, Kiwoong Park, and Victor Lempitsky.
\newblock Resolution-robust large mask inpainting with fourier convolutions.
\newblock \emph{arXiv preprint arXiv:2109.07161}, 2021.

\bibitem[Wang et~al.(2023)Wang, Zhao, and Xing]{wang2023stylediffusion}
Zhizhong Wang, Lei Zhao, and Wei Xing.
\newblock Stylediffusion: Controllable disentangled style transfer via diffusion models, 2023.

\bibitem[Wu et~al.(2015)Wu, Song, Khosla, Yu, Zhang, Tang, and Xiao]{wu20153d}
Zhirong Wu, Shuran Song, Aditya Khosla, Fisher Yu, Linguang Zhang, Xiaoou Tang, and Jianxiong Xiao.
\newblock 3d shapenets: A deep representation for volumetric shapes.
\newblock In \emph{Proceedings of the IEEE conference on computer vision and pattern recognition}, pages 1912--1920, 2015.

\bibitem[Yu et~al.(2022)Yu, Xu, Koh, Luong, Baid, Wang, Vasudevan, Ku, Yang, Ayan, Hutchinson, Han, Parekh, Li, Zhang, Baldridge, and Wu]{parti}
Jiahui Yu, Yuanzhong Xu, Jing~Yu Koh, Thang Luong, Gunjan Baid, Zirui Wang, Vijay Vasudevan, Alexander Ku, Yinfei Yang, Burcu~Karagol Ayan, Ben Hutchinson, Wei Han, Zarana Parekh, Xin Li, Han Zhang, Jason Baldridge, and Yonghui Wu.
\newblock Scaling autoregressive models for content-rich text-to-image generation.
\newblock \emph{Transactions on Machine Learning Research}, 2022.

\bibitem[Zeng et~al.(2023)Zeng, Lin, Zhang, Liu, Collomosse, Kuen, and Patel]{zeng2022scenecomposer}
Yu Zeng, Zhe Lin, Jianming Zhang, Qing Liu, John Collomosse, Jason Kuen, and M. Patel, Vishal.
\newblock Scenecomposer: Any-level semantic image synthesis.
\newblock 2023.

\bibitem[Zhang et~al.(2023)Zhang, Rao, and Agrawala]{zhang2023_controlnet}
Lvmin Zhang, Anyi Rao, and Maneesh Agrawala.
\newblock Adding conditional control to text-to-image diffusion models.
\newblock In \emph{Proceedings of the IEEE/CVF International Conference on Computer Vision}, pages 3836--3847, 2023.

\bibitem[Zhang et~al.(2018)Zhang, Isola, Efros, Shechtman, and Wang]{zhang2018unreasonable}
Richard Zhang, Phillip Isola, Alexei~A Efros, Eli Shechtman, and Oliver Wang.
\newblock The unreasonable effectiveness of deep features as a perceptual metric.
\newblock In \emph{Proceedings of the IEEE conference on computer vision and pattern recognition}, pages 586--595, 2018.

\bibitem[Zheng et~al.(2012)Zheng, Chen, Cheng, Zhou, Hu, and Mitra]{interactiveimages}
Youyi Zheng, Xiang Chen, Ming-Ming Cheng, Kun Zhou, Shi-Min Hu, and Niloy~J. Mitra.
\newblock Interactive images: Cuboid proxies for smart image manipulation.
\newblock \emph{ACM Transactions on Computer Graphics}, 2012.

\end{thebibliography}
}

\clearpage
\setcounter{page}{1}
\setcounter{section}{0}
\setcounter{figure}{0}
\setcounter{table}{0}
\renewcommand\thefigure{S\arabic{figure}}
\renewcommand\thesection{S\arabic{section}}
\renewcommand\thetable{S\arabic{table}}
\maketitlesupplementary

\section{Overview}
In the supplementary material, we provide a full quantitative comparison to all baselines on our \texttt{Benchmark} dataset in Section~\ref{sec:supp_quantitative}, give full qualitative comparison on both the \texttt{PhotoGen} and the \texttt{Benchmark} datasets in Sections~\ref{sec:supp_qual_photogen} and~\ref{sec:supp_qual_benchmark}, provide information about the diffusion sampler in Section~\ref{sec:supp_diff_sampler}, and additional information about our user study in Section~\ref{sec:supp_user_study}.

The website \href{https://diffusionhandles.github.io/}{diffusionhandles.github.io} for our method provides an overview, more detailed stop-motion results, and also hosts the full qualitative results is available in the supplementary material.

\section{Full Quantitative Comparison}
\label{sec:supp_quantitative}

We provide a full quantitative comparison on the \texttt{Benchmark} dataset, using the same metrics described in the main paper:

\noindent
(A) 
For \emph{Identity Preservation}, we use a cycle consistency metric that measures the difference between the original image and the edited image transformed back to the original object configuration using the inverse 3D transform. Denoting the target 3D edit as $T$ and the image edit performed by our method or a baseline as $\mathcal{E}$ (lower is better):
\begin{align}
    E_\text{id}^{\text{L1}} &= \|x_0 - \mathcal{E}(T^{-1}, \mathcal{E}(T, x_0)) \|_1, \text{ and }\\
    E_\text{id}^{\text{LPIPS}} &= \text{LPIPS}\big(x_0,\ \mathcal{E}(T^{-1}, \mathcal{E}(T, x_0))\big),
\end{align}
where we use either $L1$ or LPIPS~\cite{zhang2018unreasonable} to measure the image difference.

\noindent
(B) 
For \emph{Edit Adherence}, we measure the Intersection over Union (IoU) between the mask $M_e$ of the edited foreground object and the corresponding ground truth mask $M_e^\text{gt}$ (higher is better):
\begin{equation}
    S_{\text{edit}} = \text{IoU}(M_e, M^\text{gt}_e).
\end{equation}
The mask $M_e$ is using the same foreground segmentation approach described in Section~\ref{sec:3d_edit} (Step 2) that is based on SAM~\cite{kirillov2023segany}, starting from a Grounding DINO~\cite{liu2023grounding} bounding box. The ground truth mask $M^\text{gt}_e$ is obtained by applying the target 3D edit to the synthetic scenes our \texttt{Benchmark} dataset was created from. 

Results are shown in Table~\ref{tab:quant_comparison}. For additional insight, compare to the qualitative results given in Section~\ref{sec:supp_qual_benchmark}. We can see that Zero123 performs worst on identity preservation, as it introduces significant distortions during object edits that accumulate in the edit cycle. ObjectStitch suffers less from distortions, but has a lower degree of identity preservation in each edit. 3DIT performs second best, although looking at the qualitative results, we can see that this good quantitative performance is deceptive: 3DIT often fails to change the input image at all, resulting in good identity preservation, but bad edit adherence. For fairness, we ignore 3DIT results that do not change the foreground object at all when computing identity preservation. All baselines have relatively low edit adherence. ObjectStitch does not provide 3D controls, while Zero123 and 3DIT lack accuracy in their 3D controls. Our 3D-aware guidance provides both more accurate control and better identity preservation.

\begin{table}[t]
\centering
\caption{
\textbf{Quantitative comparison on the \texttt{Benchmark} dataset.} We compare \emph{identity preservation}, based on the cycle consistency of performing the edit, followed by its inverse; and \emph{edit adherence}, as measured by the IoU between image region covered by the edited foreground object and the corresponding ground truth image region.
}
\footnotesize 
\renewcommand{\arraystretch}{1.1}
\setlength{\tabcolsep}{5pt}
\begin{tabularx}{\linewidth}{r >{\centering\arraybackslash}X >{\centering\arraybackslash}X >{\centering\arraybackslash}X} 
\toprule
     
     & \multicolumn{2}{c}{Identity Preservation} & Edit Adherence \\
     \cmidrule(l    r){2-3} \cmidrule(l){4-4}
     & $E_\text{id}^{\text{L1}}$($\times 10$)$\downarrow$ & $E_\text{id}^{\text{LPIPS}}\downarrow$ & $S_{\text{edit}}\uparrow$ \\
     \midrule
     Obj.Stitch~\cite{song2023objectstitch} & 0.89 & \underline{0.25} & 0.37 \\
     Zero123~\cite{zero123}                 & 1.05 & 0.31 & \underline{0.52} \\
     3DIT~\cite{michel2023object}           & \underline{0.74} & 0.27 &  0.15 \\
     \textbf{Ours}                          & \textbf{0.71} & \textbf{0.19} & \textbf{0.85} \\
     \bottomrule
\end{tabularx}
\label{tab:quant_comparison}
\end{table}

\section{Qual. Comparison on \texttt{PhotoGen} dataset}
\label{sec:supp_qual_photogen}
The \href{https://diffusionhandles.github.io/static/full-results/qualitative_comparison_main.html}{full qualitative comparison} on all samples from the \texttt{PhotoGen} dataset is linked from our website \href{https://diffusionhandles.github.io/}{diffusionhandles.github.io}.

\section{Qual. Comparison on \texttt{Benchmark} dataset}
\label{sec:supp_qual_benchmark}
The \href{https://diffusionhandles.github.io/static/full-results/qualitative_comparison_benchmark.html}{full qualitative comparison} on all samples from the \texttt{Benchmark} dataset is is linked from our website \href{https://diffusionhandles.github.io/}{diffusionhandles.github.io}.

\begin{figure*}[t]
    \centering
    \includegraphics[width=\textwidth]{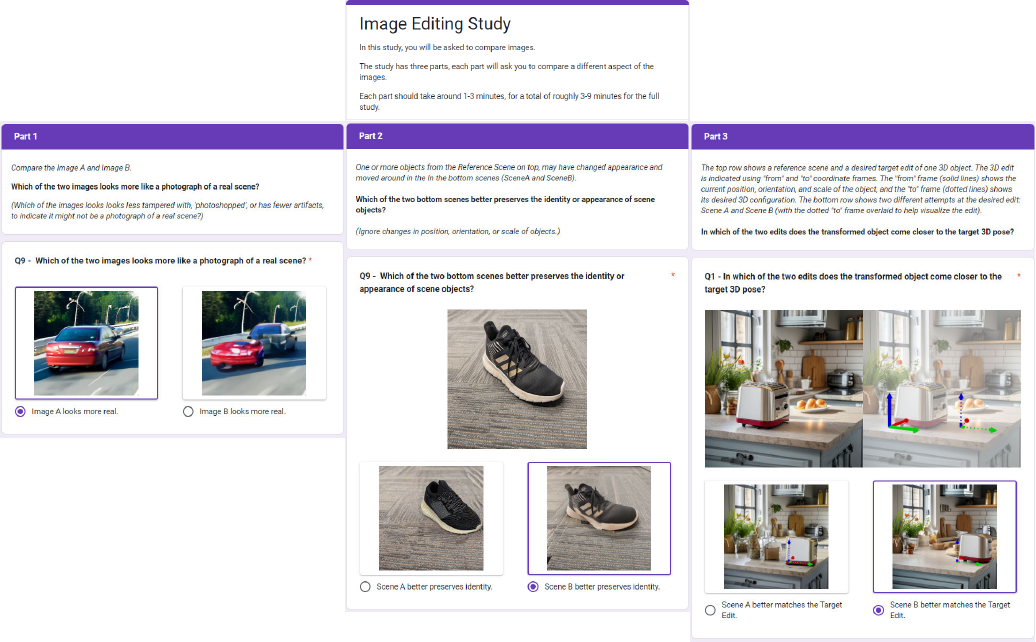}
    \caption{\textbf{User Study Screenshot.} We asked three types of questions in the user study to measure plausibility (left), identity preservation (middle), and edit adherence (right). In each question, the user chooses between two images.}
    \label{fig:supp_user_study_screenshot}
    \vspace{-5pt}
\end{figure*}

\section{Diffusion Sampler Details}
\label{sec:supp_diff_sampler}
We use the DDIM sampler as described in Denoising Diffusion Implicit Models~\cite{song2020denoising} in in all our experiments, using 50 denoising steps. When computing the edited image with our guidance energy $\mathcal{G}$, in each iteration we perform three steps of gradient descent $\nabla_{\tilde{x}(t)}\ \mathcal{G}(\tilde{x}(t); t, y, d)$ (see Eq.~\ref{eq:guidance}) on our energy to nudge the denoising trajectory in a direction that minimizes the guidance energy.

\section{User Study Details}
\label{sec:supp_user_study}

Figure~\ref{fig:supp_user_study_screenshot} shows screenshots with examples for each type of question in our user study. We split the study into three parts, corresponding to plausibility, identity preservation, and edit adherence. In each part we ask one specific type of question, and the user chooses between two images as response. We start by asking about image plausibility in the first part, as this does not require introducing the notion of a 3D edit. In the second part, we ask about identity preservation, using the input image as reference, and in the third part, we ask about edit adherence, using both the input image and a visualization of the target edit as reference.

\end{document}